\documentclass[12pt,journal,compsoc,onecolumn]{IEEEtran}
 \usepackage{amsmath,graphicx}
\usepackage{color}
 \usepackage{amssymb}
\usepackage{amsfonts}
\usepackage{dsfont}
\usepackage{epsfig}
 \usepackage{times}
\usepackage{graphicx}
\usepackage{adjustbox}
\usepackage{slashbox}
\usepackage{nicefrac}       
\usepackage{microtype}       
\usepackage{graphicx}
\usepackage{multirow}

 \usepackage{pifont}
\usepackage{lipsum}

 \usepackage{bbm}
 \usepackage{algorithmic}
\usepackage[noadjust]{cite}
 \newtheorem{definition}{Definition}

\ifCLASSINFOpdf
\else
\fi
\ifCLASSOPTIONcompsoc
 \usepackage[font=normalsize,labelfont=sf,textfont=sf]{subfig}
\else
 \usepackage[font=footnotesize]{subfig}
\fi
\ifCLASSOPTIONcompsoc
 \usepackage[font=normalsize,labelfont=sf,textfont=sf]{subfig}
\else
 \usepackage[font=footnotesize]{subfig}
\fi

 \def\A{{\bf A}}
\def\Ak{{\bf A}^k}
\def\I{{\bf I}}
\def\X{{\bf X}}
\def\B{{\bf B}}
\def\K{{\bf K}}
\def\U{{\bf U}}
\def\W{{\bf W}}

\def\S{{\cal S}}
\def\N{{\cal N}}

\def\G{{\cal G}}
\def\V{{\cal V}}
\def\E{{\cal E}}
\def\F{{\cal F}}
\def\z{{\bf z}}
\def\K{{\cal K}}

\def\x{{\bf x}}
\def\y{{\bf y}}
\def\betaa{{\hat{\w}}}

\begin{document}

\title{Action Recognition with Kernel-based Graph Convolutional Networks} 
\author{Hichem Sahbi\\
Sorbonne University, CNRS, LIP6\\
F-75005, Paris, France 
}

\maketitle

\begin{abstract}
  Learning graph convolutional networks (GCNs) is an emerging field which aims at generalizing deep learning to arbitrary non-regular domains. Most of the existing GCNs follow a neighborhood aggregation scheme, where the representation of a node is recursively obtained by aggregating its neighboring node representations using averaging or sorting operations. However, these operations are either ill-posed or weak to be discriminant or increase the number of training parameters and thereby the computational complexity and the risk of overfitting. \\ 
  In this paper, we introduce a novel GCN framework that achieves spatial graph convolution in a reproducing kernel Hilbert space (RKHS). The latter makes it possible to design, via implicit kernel representations, convolutional graph filters in a high dimensional and more discriminating space without increasing the number of training parameters. The particularity of our GCN model also resides in its ability to achieve convolutions without explicitly realigning nodes in the receptive fields of the learned graph filters with those of the input graphs, thereby making convolutions permutation agnostic and well defined.  Experiments conducted on the challenging task of skeleton-based action recognition show the superiority of the proposed method against different baselines as well as the related work.
\end{abstract}

 \section{Introduction}
There is an increasing interest in deep learning for different pattern classification and recognition tasks \cite{sahbiicassp2015,sahbiicassp2019,sahbiacm2000}. These parametric models rely on deep neural networks, composed of several convolutional, pooling and fully connected layers,  that capture different levels of abstractions in the analyzed patterns \cite{goodfellow2016deep}. These models have been popular in the analysis of vectorial data; i.e., those sitting on top of regular domains such as images \cite{alexnet2012,inception2015,sahbicassp2016a,sahbiccv2017,sahbiPR2019,resnet2016,mobilenet2017,squeezenet2016}. However, the extension of these models to non-regular domains, such as graphs, remains a major challenge even though interesting solutions are currently emerging~\cite{kipf2016semi,bruna2013spectral,sahbiBMVC2019,defferrard2016convolutional,gao2018large,huang2018adaptive,monti2017geometric}. Indeed, the difficulty in analyzing non-vectorial data stems from the ambiguity in defining usual operations namely convolutions. Whereas achieving convolution using sliding windows in regular domains, such as images, is a well defined operation, there is no clear definition of sliding windows in general graphs \cite{monti2017geometric}; besides, the number and the order of nodes that intervene in the receptive fields of convolutions may change dramatically across different graph instances. \\
\indent Early graph convolutional network (GCN) methods \cite{gori2005new,micheli2009neural,scarselli2008graph,wu2019comprehensive} and their variants (see for instance \cite{li2015gated,dai2018learning,bacciu2018,zhang2018,ying2018,genie2019,xu2018}) are rather spatial and seek to learn graph representations by iteratively propagating node features (a.k.a representations, descriptions or signals) through their neighbors using recurrent neural architectures till a stationary point is reached. These spatial methods also include recurrent gaited networks \cite{scarselli2008graph,li2015gated,dai2018learning} that share the same convolutional parameters through layers, and composition-based convolutional networks \cite{hamilton2017inductive} that consider different parameters. However, on highly irregular graphs, convolutions are ill-posed as the notion of translation and filter support (i.e., receptive field) cannot be consistently defined. Existing attempts, to address these issues,  achieve node sorting and efficient sampling of neighboring nodes in order to define the receptive field during graph convolutions  \cite{chen2017stochastic}  and to make it similar to regular (grid-like) domains \cite{atwood2016diffusion,niepert2016learning,gao2018large}. Other solutions operate differently \cite{hamilton2017inductive,monti2017geometric,niepert2016learning,gao2018large,zhang2018gaan}; first, they describe nodes by aggregating their neighbors into fixed length features prior to apply convolution (based on inner product) on the aggregated features. \\ 
\noindent On another hand, spectral methods provide interesting alternatives to make convolutions well defined \cite{bruna2013spectral,defferrard2016convolutional,kipf2016semi,henaff2015deep,li2018adaptive,levie2018cayleynets,dual2018}. These methods rely on the Fourier transform that projects the signal of a given graph using the spectral decomposition of its Laplacian prior to perform convolution in the Fourier domain, and then back-project the result in the input domain; in particular, the method in \cite{defferrard2016convolutional} makes it possible to project graph signals using an orthogonal Chebyshev basis prior to achieve convolution. An extension, in \cite{kipf2016semi}, allows to reduce the Chebychev polynomial using a first order approximation which provides a spatially localized convolution, that is equivalent to spatial methods. A variant  in \cite{chen2018fastgcn}  interprets the graph convolutions in \cite{kipf2016semi} as integral transforms of embedding functions under probability measures and uses Monte Carlo sampling to efficiently and consistently estimate the integrals. Huang et al. \cite{huang2018adaptive} propose an adaptive layer-wise sampling approach, based on variance reduction in order to accelerate the training of ChebyshevNet~\cite{kipf2016semi}, where sampling for a lower layer is conditioned on a top one. Nonetheless, most of these spectral methods suffer from several drawbacks; the eigen decomposition of the Laplacian, besides being computationally expensive, is sensitive to any small perturbation of input graphs (that may result from the intra-class variability). Moreover, the learned filters are domain dependent and cannot be transferred to graphs with high topological variations.\\
\noindent Besides the aforementioned issues, the accuracy of both spatial and spectral GCNs also relies on the discrimination power of the input graph signal. For highly nonlinear graph signals, relying on convolutions in the input space may limit the discrimination power of the learned convolutional representations and may result into limited accuracy. Furthermore, sorting using automorphisms is not always consistent through different graph instances while aggregation based on averaging (when achieved in the input space) may dilute input node representations prior to convolution. An explicit expansion of the input node representations may enhance the discrimination power but comes at the expense of a substantial increase in the number of training parameters (thereby the risk of overfitting) and also an increase in the computational complexity both in space and time. Therefore, one should consider, instead, an {\it implicit} mapping of the input graph signal in a (high or possibly infinite dimensional) reproducing kernel Hilbert space ({\it RKHS})~\cite{vapnik1998} and achieve an averaging aggregation and convolution in that space, in order to enhance the representational power of nodes and also the learned graph representations while being permutation agnostic. This mapping scheme has been proven to be effective in kernel methods, and particularly in support vector machines (SVMs) (see for instance~\cite{sahbirr2002,kernels2004,sahbirr2004,sahbivisual2004,sahbisoft2008,sahbitnnls2017,sahbijstars2017,sahbiccv2013,sahbineuro2007,sahbicip2014,sahbipami2011}) and it is extended, in our paper, to GCNs.\\
\indent Considering these challenges, we introduce in this paper a {\it dual} formulation of GCN based on kernels which maps graph signals from an input space into a high dimensional Hilbert space. This mapping is implicitly defined using positive semi-definite kernels that enhance the discrimination power of the learned graph  representations, without explicitly increasing the dimensionality of the input graph signals nor the number of training parameters\footnote{In contrast to \cite{hamid2014compact,kar2012random,le2013fastfood,vedaldi2012efficient,rahimi2008random,dai2014scalable} which may increase the number of parameters in the model and the risk of overfitting.}. This is beneficial when handling low dimensional raw signals, such as  3D skeleton graphs in action recognition~\cite{ntu2016,SBU12}; indeed the low dimensionality of these data makes the Bayes risk of the underlying classification task intrinsically high, and this requires increasing the dimensionality of the input raw signal. Moreover, our GCN   achieves convolutions without explicitly realigning nodes in the receptive fields of the learned graph filters with those of the input graphs, thereby making convolutions permutation agnostic. We cast the problem of filter design as kernel learning with the particularity of using standard kernels while training only their support vectors;  this scheme of learning the support vectors (as a part of kernel design) is conceptually different from the two major families of kernel learning techniques, namely non-parametric \cite{vo2012transductive,sahbiPR2012} and parametric ones~\cite{gonen2011multiple,cortes2009learning,sahbiTIP2017}\footnote{In "non-parametric" training, the number of parameters follows exactly the size of training data (e.g.,  nonlinear SVMs) while in  "parametric" training, this number is fixed independently  (e.g., linear SVMs). In "semi-parametric" training, only a fraction of the parameters follows proportionally the size of training data.}.  Finally, extensive experiments on the challenging task of action recognition show the high gain of our kernel-based GCNs w.r.t standard baselines as well as the related work. 

\def\A{{\bf A}}
\def\Ak{{\bf A}^k}
\def\I{{\bf I}}
\def\X{{\bf X}}
\def\B{{\bf B}}
\def\K{{\bf K}}
\def\U{{\bf U}}
\def\W{{\bf W}}

\def\S{{\cal S}}
\def\N{{\cal N}}

\def\G{{\cal G}}
\def\V{{\cal V}}
\def\E{{\cal E}}
\def\F{{\cal F}}
\def\z{{\bf z}}

\section{Graph convolutional networks} 

Let $\S=\{\G_i=(\V_i,\E_i)\}_i$ denote a collection of graphs with $\V_i$, $\E_i$ being respectively the nodes and the edges of $\G_i$. Each graph $\G_i$ (denoted for short as $\G=(\V,\E)$) is endowed with (i) a signal $\{s(u) \in {\cal X}: \ u \in \V\}$ (with ${\cal X}=\mathbb{R}^D$ being an input space) and (ii) a row-stochastic  adjacency matrix $\A$ with each entry  $\A_{uu'}>0$ iff $(u,u') \in \E$ and $0$ otherwise. Our goal is to design a novel graph convolutional network that returns both the representation and the classification of $\G$.  

\def\K{{\cal K}}

\subsection{Standard graph convolutional networks}\label{standardGCN}
Consider $\G=(\V,\E)$, $g_\theta=(\V_\theta,\G_\theta)$ as two graphs with $|\V_\theta| \ll |\V|$ and $|\E_\theta| \ll |\E|$. Following standard GCNs (see for instance~\cite{monti2017geometric}), the spatial convolution of $\G$ with a graph $g_\theta$ at a given node $u \in \V$ is defined as \begin{equation}\label{initial0} 
  (\G \star g_\theta)_u = \sigma( \K_\theta(u)), 
\end{equation}
with 
\begin{equation}\label{initial0} 
  \K_\theta(u) =  \bigg \langle \sum_{u'}  s(u') . [\A^r]_{uu'}, w_\theta \bigg \rangle, 
\end{equation}
here $\sigma(.)$ is a nonlinear activation (taken in practice as ReLU), $w_\theta \in {\cal X}$ corresponds to the filter parameters of the graph $g_\theta$ (also referred to as graphlet) and  $[\A^r]_{uu'}$ is the ${u'}^{th}$  column of the ${u}^{th}$  row of the $r$-hop adjacency matrix $\A^r$. In this definition, the left-hand side term of the inner product in Eq.~\ref{initial0}, aggregates the neighbors of $u$ into a single vector prior to multiply this vector by $w_\theta$. \\
In spite of being agnostic to any arbitrary permutation of nodes in $\G$, the above definition suffers from limited discrimination power, as the signal informations in the neighborhood system $\{\N_r(u)\}_u$ of $\G$ are mixed during convolution. In what follows, we consider a dual convolutional operator, based on kernels,  that overcomes this limitation and provides more discriminating convolutional representations while still being agnostic to any arbitrary permutation of nodes in graphs.

\subsection{Our kernel-based graph convolutional networks}

\indent Considering $\kappa$ as a symmetric positive definite function (i.e., $\exists \psi: {\cal X} \rightarrow {\cal H}$, with $\psi$ being an implicit mapping that takes graph signals from an input space $\cal X$  to a high dimensional Hilbert space $\cal H$, s.t.,  $\kappa(s(u'),s(v))=\langle \psi(s(u')), \psi(s(v))\rangle$) and for a particular setting of $w_\theta$ as $\frac{1}{|\V_\theta|}\sum_{i=1}^N \alpha_i^\theta \psi(s(v_i^\theta))$\footnote{This setting is related to the representer theorem widely used in kernel methods~\cite{representer2001,wahba1971}. The latter states that many problems have optimal solutions that live in a finite dimensional span of training data mapped into a high dimensional Hilbert space, and this makes it possible to define  kernel-based algorithms independently of the (high or infinite) dimensionality of these Hilbert spaces.}, with $\{v_i^\theta\}_i \subset \V_\theta$, $\{\alpha_i^\theta\}_i \subset \mathbb{R}$; the convolutional operator defined in Eq.~\ref{initial0} can be rewritten as
\begin{equation}\label{initial} 
\K_\theta(u)= \frac{1}{|\N_r(u)|.|\V_\theta|} \sum_{u' \in \N_r(u)} \bigg(\sum_{i=1}^N \alpha_i^\theta \  \kappa(u',v_i^\theta)\bigg),
\end{equation}
 
here $\N_r(u)$ is the set of $r$-hop neighbors of $u$ and  $\kappa(s(.),s(.))$, $\psi(s(.))$ are written for short as $\kappa(.,.)$ and $\psi(.)$ respectively. In the above definition, 
$\{v_i^\theta\}_{i,\theta} $ are referred to as support vectors and $\alpha=\{\alpha_i^\theta\}_{i,\theta} $ as the underlying mixing parameters.  Since  $\K_\theta(u)$ is defined as the sum of all of the kernel values between all of the possible signal pairs taken from $\N_r(u) \times \V_\theta$, its evaluation does not require  any explicit alignment between these pairs and it is thereby still invariant to any arbitrary permutation (including rotations) of nodes in $\V$ and $\V_\theta$. \\
\indent The strength of this kernel trick resides in its  capacity to handle nonlinear data as node representations are mapped into a high dimensional (and more discriminating) space ${\cal H}=\mathbb{R}^H$. For instance, when using the polynomial kernel $\kappa(s(u),s(v)) = \langle s(u),s(v)\rangle^p$, its underlying mapping is explicitly defined as $\psi(s(u))=s(u) \otimes \dots \otimes  s(u)$ (with $\otimes$ being the Kronecker tensor product applied $p-1$ times); see also \cite{sahbiijmir2016,maji2012efficient,vedaldi2012efficient}. As the dimensionality $H$ of this explicit map grows exponentially w.r.t $p$ and polynomially w.r.t  $D$, the kernel form is rather computationally more efficient. Indeed, considering a non-parametric setting with a fixed  set of  support vectors $\{v_i^\theta\}_i$  taken from the training set (i.e., $\cup \ \V_j$);  when only $\{\alpha_i^\theta\}_i$ are allowed to vary in  $w_\theta=\frac{1}{|\V_\theta|}\sum_{i} \alpha_i^\theta \psi(v_i^\theta)$, and when $H \gg |\{\alpha_i^\theta\}_i|$, the kernel trick presented earlier provides a computational and a generalization advantage (i.e., the convolution in Eq.~\ref{initial} has fewer parameters compared to the one in Eq.~\ref{initial0}).  However, this may still come at the expense of a {\it quadratic} complexity when naively evaluating $\{\kappa(.,.)\}$; for mid (and even small) scale training problems with a large number of nodes in $\cup \ \V_j$, this complexity becomes clearly intractable. \\ 
\noindent One question that arises is how to make this approach parametric (or at least semi-parametric); in other words, how to maintain the kernel trick advantage (in Eq.~\ref{initial}) without significantly increasing the computational cost w.r.t   the total number of nodes in $\cup \ \V_j$. Solutions such as sampling and reduced set technique \cite{burges1997improving} are both limited; on the one hand, sampling may generate a smaller fixed set of support vectors $\{v_i^\theta\}_i$ but biased (i.e., very limited to comprehensively make $w_\theta$ a  universal filter approximator). On the other hand,  the reduced set technique requires first building an initial expensive model $w_\theta$ before  reducing its complexity by solving a difficult pre-image optimization problem \cite{burges1997improving,sahbiphd2003,sahbijmlr2006}. Our alternative, in this work, is to control the size  of $\{v_i^\theta\}_i$  while allowing  entries in $\{v_i^\theta\}_i$ to vary as a part of the end-to-end GCN (and also kernel) learning; this makes it possible to model a larger class of filters $\{w_\theta\}$ that better fit the classification task at hand (see later experiments). \\ 
\indent Note that one may consider a kernel approximation $\hat{\psi}(.)$ s.t. $\kappa(.,.)\approx \langle \hat{\psi}(.),\hat{\psi}(.) \rangle$ (as done in  \cite{lu2014scale,sahbijiuicassp2016,hamid2014compact,kar2012random,le2013fastfood,vedaldi2012efficient,rahimi2008random,dai2014scalable,cho2009kernel} which seek to handcraft or learn shallow/deep explicit maps whose inner products approximate the original kernel values) and use instead Eq.~\ref{initial0}. However, this approximation usually results into very high dimensional mappings (and hence into a lot of training parameters in $w_\theta$), especially when considering highly nonlinear (and also discriminative) kernels  such as gaussian, histogram intersection and triangular \cite{sahbistcv2003,sahbiicip2002}. Put differently, even when learning both  $\{v_i^\theta\}_i$ and $\{\alpha_i^\theta\}_i$, the dual formulation in Eq.~\ref{initial} is computationally more efficient and less subject to overfitting, as the dimensionality $H$ of $\hat{\psi}$ is often $\gg |\V_\theta| \times D$ for the widely used kernels including  gaussian and histogram intersection. In sum, our method is rather targeted to learn kernels following a (semi-)parametric setting by allowing the support vectors of these kernels to be learned (instead of being taken from training data) and this is also conceptually very different from multiple kernel learning \cite{gonen2011multiple}.

\subsection{Neural consistency and architecture design}    
In contrast to usual convolutional operators on graphs (including  Eq.~\ref{initial0}), the one in Eq.~\ref{initial} cannot be straightforwardly evaluated using standard neural units\footnote{i.e., those based on standard perceptron (inner product operators) followed by nonlinear activations.} as kernels may have general forms. Hence, modeling Eq.~\ref{initial} requires a careful design; our goal in this paper, is not to change the definition of neural units, but instead to adapt Eq.~\ref{initial} in order to make it consistent with the usual definition of neural units. In what follows, we introduce  the overall architecture associated to $\K_\theta(.)$ (and the whole GCN) for different kernels including linear, polynomial, gaussian and histogram intersection as well as a more general class of shift invariant kernels.\\  
\def\x{{\bf x}}
\def\y{{\bf y}}

\begin{definition}[Neural consistency] Let $u_{.,d}$ (resp. $v_{.,d}$) denote the $d^{th}$ dimension of the signal in a given node $u$ (resp. $v$). For a given (fixed or learned) $v$, a kernel $\kappa$ is referred to as ``neural-consistent'' if
\begin{equation}\label{expansion0} 
  \kappa(u,v) = \sigma_3\bigg(\sum_d \sigma_2(\sigma_1(u_{.,d}).\omega_d)\bigg),
  \end{equation} 
with   $\omega_d=\sigma_4(v_{.,d})$ and being  $\sigma_1$, $\sigma_2$, $\sigma_3$, $\sigma_4$ any arbitrary real-valued activation functions.\label{define000}   
\end{definition}

Considering the above definition, the following kernels are neural consistent: linear $\langle u,v\rangle$, polynomial $\langle u,v\rangle^p$, and $\tanh(a \langle u,v\rangle+b)$. Neural consistency is straightforward for inner product-based kernels (namely linear, polynomial and tanh) while for shift-invariant ones such as the gaussian, one may obtain neural consistency by rewriting $\exp(-\beta \|u-v\|_2^2) = \sigma_3\big(\sum_d \sigma_2(\sigma_1(u_{.,d}).\omega_d)\big)$ with $\sigma_1(.)=\exp(.)$, $\sigma_2(.) = \log(.)^2$, $\sigma_3(.)= \exp(-\beta (.))$ and $\omega_d=\exp(-v_{.,d})$. Other kernels (including Laplacian, inverse multiquadric, power, log, Cauchy\footnote{See for instance \cite{genton2001classes} for a taxonomy of the widely used functions in kernel machines.}) are also neural consistent (see table~\ref{taxi} for the setting of their $\sigma_1$, $\sigma_2$, $\sigma_3$, $\sigma_4$). \\ For the histogram intersection kernel, $\sum_d \min(u_{.,d},v_{.,d})  =  \sum_d 1-\max(1-u_{.,d},1-v_{.,d})$ and one may easily obtain $\sum_d 1-\max(1-u_{.,d},1-v_{.,d}) \approx \sigma_3\big(\sum_d \sigma_2(\sigma_1(u_{.,d}).\omega_d)\big)$ using $\sigma_1(.)=\exp(\exp(\beta(1-(.))))$, $\sigma_2(.) = -\frac{1}{\beta}\log(\log(.))+1$, $\sigma_3(.)=(.)$ and $\omega_d=\sigma_1(v_{.,d})$ (for a sufficiently large $\beta$).  In the following section, we discuss the implementation details of our global GCN architecture built on top of these neural consistent kernels. 
\begin{table*}
  \begin{center}
    \resizebox{1.02\textwidth}{!}{
  \begin{tabular}{cc||c|cccc}
      & & $\kappa(u,v)$ &  $\sigma_1(t)$ & $\sigma_2(t)$ & $\sigma_3(t)$ & $\sigma_4(t)$  \\
    \hline
    \hline 
     \multirow{4}{*}{\rotatebox{38}{\scriptsize Inner product based}} &  Linear  & $\langle u,v \rangle$ & $t$  & $t$  & $t$  & $t$    \\   
    & Polynomial  &$\langle u,v \rangle^p$ & $t$ &  $t$  & $t^p$ & $t$  \\   
    &Sigmoid & $\frac{1}{1+\exp(-\beta \langle u,v\rangle )}$ & $t$ & $t$  & $\frac{1}{1+\exp(-\beta t)}$ &  $t$  \\  
    &tanh  &$\tanh (a \langle u,v \rangle + b$) & $t$ & $t$  & $\tanh(a t + b)$ &  $t$  \\   

  \hline
    \multirow{7}{*}{\rotatebox{38}{\scriptsize Distance based}}  &  Gaussian  & $\exp(-\beta \|u-v \|^2)$ & $\exp(t)$  & $\log(t)^2$& $\exp(-\beta t)$ & $\exp(-t)$  \\   
   & Laplacian  &$\exp(-\beta \|u-v \|)$& $\exp(t)$  & $\log(t)^2$& $\exp(-\beta \sqrt{t})$ & $\exp(-t)$  \\  
    &Power & $-\|u-v\|^p$ &$\exp(t)$  & $\log(t)^2$& $-t^{p/2}$ & $\exp(-t)$  \\  
  
    & Inverse Multi-quadric  &$\frac{1}{\sqrt{\|u-v\|^2+b^2}}$ & $\exp(t)$  & $\log(t)^2$ & $\frac{1}{\sqrt{t+b^2}}$  &   $\exp(-t)$  \\   
    &Log & $-\log(\|u-v\|^p+1)$  & $\exp(t)$  & $\log(t)^2$ & $-\log(t^{p/2}+1)$ &  $\exp(-t)$     \\   
    & Cauchy  & $\frac{1}{1+\frac{\|u-v\|^2}{\sigma^2}}$  & $\exp(t)$  & $\log(t)^2$ & $\frac{1}{1+\frac{t}{\sigma^2}}$ &  $\exp(-t)$   \\  
    \hline
      &  Histogram intersection & $\sum_d \min(u_{.,d},v_{.,d})$ & $\exp(\exp(\beta(1-t)))$& $-\frac{1}{\beta} \log(\log(t))+1$ & $t$  & $\sigma_1(t)$

  \end{tabular}}
\end{center}
\caption{This table shows the setting of $\sigma_1$, $\sigma_2$, $\sigma_3$, $\sigma_4$ for different kernel functions.} \label{taxi}
\end{table*} 
\begin{figure*}[hpbt]
\centering \resizebox{0.95\textwidth}{!}{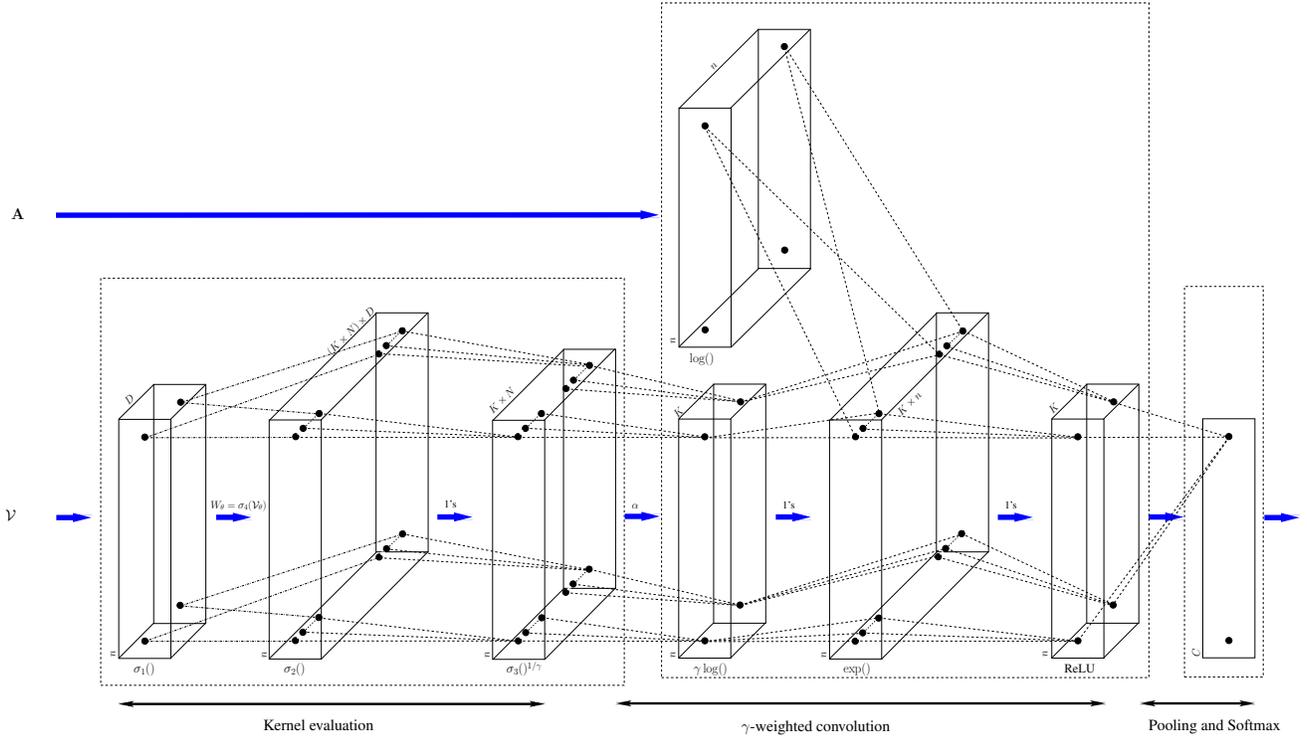}
  
\caption{This figure shows the architecture of our kernel-based graph convolutional network; see a detailed description of this architecture in the ``Implementation Part'' of section~\ref{impl} {\bf (better to zoom the PDF version).}} \label{fig01}
\end{figure*}
\subsubsection*{\bf Implementation}\label{impl}
Fig.~\ref{fig01} shows the architecture of our deep net including kernel evaluation and the weighted convolution blocks. The former block is fed with the input graph signal $s(\V)$ (denoted for short as $\V$) and the adjacency matrix $\A$ following the same arbitrary order both in $\V$ and $\A$. In the first layer, the $\sigma_1$ activation is first applied to all the dimensions of the signal $\V$, then each dimension of the resulting activated signal $\sigma_1(\V)$ is multiplied, in the second layer, by the $K\times N$ (reparameterized) weights of the node filters $\{\sigma_4(\V_\theta)\}_\theta$ (as shown in Eqs.~\ref{initial}, ~\ref{expansion0}) prior to apply the $\sigma_2$ activation; here $K$ corresponds to the number of filters and $N$ the number of nodes in the expansion of each filter. Note that these weights are shared through different nodes in $\V$. In the third layer, the results of the previous one are pooled across dimensions resulting into $N\times K$ kernel values per node in $\V$. These kernel values are activated by   $\sigma_3()$ and  fed to the weighted convolutional block in order to evaluate their weighted linear combinations,  in the fourth layer, resulting into $K$ pooled kernel values per node (see again Eqs.~\ref{initial},~\ref{expansion0}). These pooled kernel values are crossed, in the fifth layer, with the nonzero entries of the adjacency matrix $\A$ in order to make the receptive field of the convolutional operation local. Note also that the activation functions $\log()$ and  $\exp()$ are successively applied in the fourth and fifth layers in order to make this crossing operation neural consistent. Indeed, one may rewrite Eq.~\ref{initial} as
\begin{equation}\label{initial3}
\small   
  \begin{array}{lll}
\displaystyle     \K_\theta(u)
               &= & \displaystyle   \frac{1}{|\V_\theta|} \sum_{u'} \exp\bigg (\log \A_{uu'} + \log \sum_{i=1}^N \alpha_i^\theta \  \kappa(u',v_i^\theta)\bigg),
                    
    \end{array} 
\end{equation}
which corresponds to the neural consistent form shown in Eq.~\ref{expansion0}. The results of this fifth layer are pooled, in the sixth layer, through the neighborhood systems  $\{\N_r(u)\}_u$ and fed to the ReLU activation resulting into $K$ features per node in $\V$. Finally, these node features are used for final pooling and softmax classification.  

\section{Experimental validation}\label{section4}

We evaluate the performance of our kernel-based GCN (KGCN) on the challenging task of action recognition, using the SBU kinect dataset \cite{SBU12}. The latter is an interaction dataset acquired  using the Microsoft kinect sensor; it includes in total 282 video sequences\footnote{In contrast to other  visual analysis tasks (e.g.,\cite{sahbicip2014,sahbiigarss2012,sahbiicip2009,sahbiECCV2014,sahbicassp2013,sahbigarss11,sahbiTIP2013,sahbicip2001,sahbicisp2001}), images/videos are already processed and skeleton data are  available.}  belonging to $C=8$ categories:  ``approaching'', ``departing'', ``pushing'', ``kicking'', ``punching'', ``exchanging objects'', ``hugging'', and ``hand shaking'' with variable duration, viewpoint   changes and    interacting individuals (see examples in  Fig. \ref{fig1}). In all these experiments, we use the same evaluation protocol as the one suggested in \cite{SBU12} (i.e., train-test split) and we report the average accuracy over all the classes of actions.
\def\betaa{{\hat{\w}}}

\subsection{Video skeleton description}\label{graphc}
\begin{figure}[hpbt]
  \begin{center}
    \centerline{\scalebox{0.29}{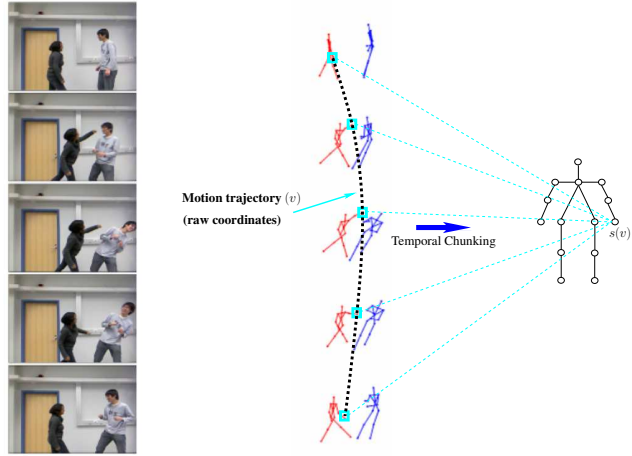}}
\vspace{-0.75cm}\caption{ This figure shows the whole keypoint  tracking and description process.} \label{fig1}
\end{center}
\end{figure}

\indent Given a  video $\V$ in SBU as a sequence of skeletons, each keypoint in these skeletons defines a labeled trajectory through successive frames (see Fig.~\ref{fig1}).   Considering a finite collection of trajectories $\{v_j\}_j$ in $\V$, we process each trajectory  using {\it temporal chunking}: first we split the total duration of a  video into $M$ equally-sized temporal chunks ($M=8$ in practice), then we assign  the keypoint  coordinates of  a given trajectory $v_j$  to the $M$ chunks (depending on their time stamps) prior to concatenate the averages of these chunks and this produces the description of $v_j$ (again denoted as $s(v_j) \in \mathbb{R}^{D}$ with $D=3 \times M$) and $\{s(v_j)\}_j$  constitutes the raw description of nodes in a given video $\V$. Note that two trajectories $v_j$ and $v_k$,  with similar keypoint coordinates but arranged differently in time, will be considered as very different when using temporal chunking. Note also that beside being compact and discriminant, this temporal chunking gathers advantages --  while discarding drawbacks -- of two widely used families of techniques mainly {\it global averaging techniques} (invariant but less discriminant)  and  {\it frame resampling techniques} (discriminant but less invariant). Put differently, temporal chunking produces discriminant raw descriptions that preserve the temporal structure of trajectories while being {\it frame-rate} and {\it duration} agnostic.

\subsection{Performances and comparison} 
\begin{figure}[hpbt]
  \begin{center}
    \centerline{\scalebox{0.46}{\includegraphics{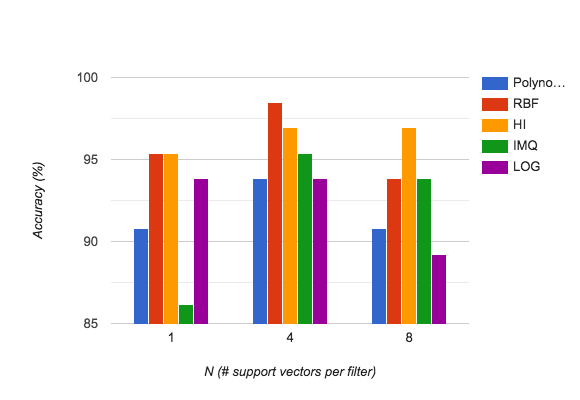}}}
  \vspace{-0.3cm}
        \centerline{\scalebox{0.46}{\includegraphics{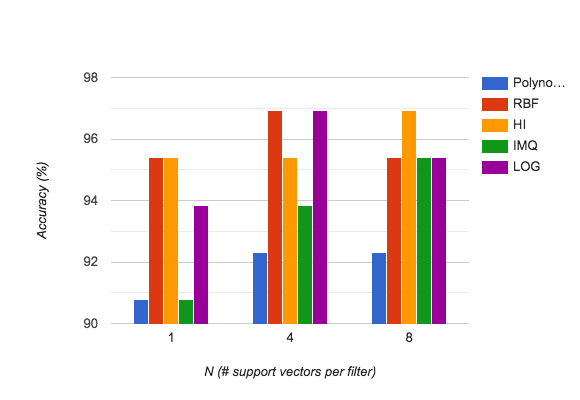}}}
\caption{Accuracy of KGCN w.r.t  five  examples of kernels and filter sizes $N$. Top figure corresponds to $K=5$ filters while bottom one to $K=10$. { \bf (Best viewed in color)}.} \label{variation11}
\end{center}
\end{figure}

We trained our kernel-based GCN end-to-end for 3000 epochs  with a batch size equal to $50$, a momentum of $0.9$ and we set the learning rate (denoted as $\nu$)  iteratively inversely proportional to the speed of change of the cross entropy loss used to train our network; when this speed increases (resp. decreases),   $\nu$  decreases as $\nu \leftarrow \nu \times 0.99$ (resp. increases as $\nu \leftarrow \nu \slash 0.99$). All these experiments are run on a GeForce GTX 1070 GPU device (with 8 GB memory) and no data augmentation is achieved.  Table~\ref{table1}  shows a comparison of action recognition performances (and also runtime per epoch during training), using our KGCN (with different kernels) against standard GCN (referred to as SGCN), shown in section~\ref{standardGCN}, with precomputed node representations based on kernel principal component analysis (KPCA) achieved on $\{s(u): u \in \cup \V_i\}$ using different kernels; in these results, we consider different numbers of eigenvectors (projection axes) corresponding to the largest eigenvalues of KPCA.\\
\begin{table*}
  \begin{center}
    \resizebox{0.99\textwidth}{!}{
  \begin{tabular}{c||cccccccccc|c}
    \backslashbox{kernels}{GCNs}  &  \multicolumn{10}{c|}{Standard GCN with different \# of KPCA dimensions ($H$) } & Our KGCN\\
                                &10 & 50  & 100& 200 & 300 & 400 & 500& 1000 & 2000& 3000  \\
    \hline
    \hline
    Linear    &   92.3077   & overdim&overdim &overdim &overdim   & overdim       & overdim         &   overdim       &  overdim         &  overdim             &     90.7692              \\
    Poly   &89.2308&95.3846& 92.3077 & 93.8462 & 93.8462 & 93.8462 & 93.8462 & overdim & overdim &  overdim &  93.8462 \\
    tanh   &  89.2308   &93.8462& 90.7692  &93.8462 & 90.7692  &  92.3077      &   93.8462       &     92.3077     &     93.8462      &    92.3077           &        96.9231           \\
    sigmoid    &93.8462 &90.7692&    93.8462    &92.3077 &  92.3077 &  92.3077      &      92.3077    &      96.9231    &     93.8462      &     92.3077          &  95.3846                 \\
     Gaussian  &92.3077&  92.3077    &92.3077 &92.3077 & 96.9231  & 93.8462       &    93.8462      &    93.8462      &    93.8462       &       93.8462        &                 98.4615  \\
         Laplacian    & 92.3077 &93.8462& 95.3846 & 92.3077 & 90.7692 & 90.7692 & 95.3846 & 93.8462 & 90.7692 & 90.7692 &  98.4615 \\
     Power  &90.7692  &92.3077&95.3846 & 92.3077  &92.3077 &95.3846  &95.3846 &93.8462  &93.8462 & 92.3077  & 96.9231  \\
     IMQ   &  87.6923&92.3077& 95.3846  &   95.3846 & 93.8462  &  93.8462         & 90.7692  & 95.3846  &  93.8462  &  93.8462  & 95.3846 \\
     Log   &   93.8462 &92.3077&  92.3077   &   95.3846 &    93.8462 &     93.8462     &     95.3846       &      90.7692     &     95.3846          &           90.7692        &          96.9231         \\
   Cauchy  & 93.8462 &95.3846&95.3846      &  92.3077&    96.9231 &  93.8462     &    92.3077      &    95.3846      &     92.3077      &       93.8462         &    98.4615               \\
   HI   & 93.8462   &92.3077&89.2308     & 90.7692 &      92.3077  &   92.3077 &    87.6923      &  87.6923        &      90.7692     & 87.6923              &  96.9231       \\          
  \hline   
    time/epoch (s)  &0.032&0.057& 0.072   &0.113  &   0.150   &  0.190         & 0.229    &    0.440   &   0.840       &               1.252      &  0.210                        
  \end{tabular}}
  \end{center}  
  \caption{ This table shows a comparison of our KGCN against SGCN (with different numbers of KPCA dimensions). Note that SGCN performances are not necessarily increasing w.r.t $H$; indeed, while more dimensions capture more statistical variance, this also increases the number of training parameters and hence the risk of overfitting. Note that for linear and polynomial kernels, the number of dimensions ($H$) cannot exceed $D$ and $D^2$  respectively with $D=24$ in practice (the Kronecker tensor product defining the map of --order 2-- polynomial kernel has $D^2$  dimensions while the map of the linear kernel has obviously $D$ dimensions.)}\label{table1}
\end{table*}
 \indent From all these results in table~\ref{table1}, we observe a clear and a consistent gain of KGCN w.r.t the linear version (i.e., KGCN with linear kernel), as well as SGCN combined with different KPCA features; we observe an increase of the accuracy of the SGCN baseline when the dimension of KPCA (again denoted as $H$) is sufficiently large (without being able to overtake KGCN for most of the kernels) and performances decrease again as the underlying number of training parameters follows $H$ and this may lead to overfitting. Besides, the average runtime per epoch, with SGCN, increases substantially when $H$ grows, as the number of training parameters in the underlying network (equal to $H \times K + C \times K$) depends on $H$  while in KGCN  the number of training parameters (equal to $(D + 1) \times N \times K + C \times K$) depends only on the dimension $D$ of the original signal despite being implicitly mapped into a high dimensional space  ${\cal H} = \mathbb{R}^{H}$. In particular,  $H  \gg N\times (D+1)$ makes KGCN clearly more efficient and  still more effective compared to SGCN  (see again table~\ref{table1} and also table~\ref{variation} and Fig.~\ref{variation11});  this performance improves further as  $N$ (the number of learned support vectors $\{ v_i^\theta\}_i$ per filter in Eq.~\ref{initial})  and $K$ (the number of convolutional filters)  reach reasonably (but not very) large values, and this results from the flexibility of the filters  which learn --- with few support vectors ---  relevant {\it representatives} of nodes in training data. These performances consistently improve for all the kernels and this is again explained by the representational power of the maps of these kernels. Moreover, the ablation study in  table~\ref{table3} shows that KGCN with learned support vectors capture better the nodes in  graph data while KGCN is clearly limited when the support vectors are fixed (and thereby biased  i.e., not sufficiently representative of the actual  distribution of the nodes, see again table~\ref{table3}); hence, learning the KGCN parameters (i.e., with learned $\alpha$  and fixed support vectors) is not enough in order to recover from this bias. In sum, the gain of our KGCN results from the {\it complementary aspects of the used (implicit) kernel maps and also the modeling capacity of our KGCN  when the support vectors of these kernels (that define the convolutional filters) are also allowed to vary}.\\

 \begin{table}[!htb]
  \begin{center}
\resizebox{0.59\linewidth}{!}{
  \begin{tabular}{c||cccc}
    \backslashbox{\# of Filters ($K$)}{\# of SVs ($N$)}  & $1$ &  $4$ & $8$ \\  
    \hline
    \hline
$1$ & 84.6154 &  84.7552 &  85.1748  \\    
$5$ &  93.1469   & {\bf 95.3846} &  92.8671    \\ 
$10$ &   92.1678  & 95.1049   &95.1049
   \end{tabular}}
\end{center}
\caption{ Average accuracy (w.r.t all the used kernels in KGCN) for different numbers ($K$) and sizes ($N$) of filters. SVs stands for support vectors.} \label{variation}
  \end{table} 
 \begin{table}
  \begin{center}
    \resizebox{0.59\columnwidth}{!}{
  \begin{tabular}{c||c|c|c}
    \backslashbox{kernels}{KGCNs}  & F-SV / L-$\alpha$ &  L-SV / F-$\alpha$ &  L-SV / L-$\alpha$ \\
 
    \hline
    \hline
 
    Linear     &   89.2308 &  90.7692 &  90.7692 \\
     Polynomial   &  84.6154 &  90.7692 & 93.8462 \\
     tanh       &  87.6923 &  90.7692 &  96.9231 \\
    Sigmoid      & 95.3846  & 95.3846  & 95.3846  \\
    Gaussian     &  84.6154 & 93.8462  &  98.4615   \\
     Laplacian       & 84.6154  & 93.8462  &  98.4615 \\
    Power           &  92.3077  &  95.3846  &  96.9231 \\ 
    I. Multi-quadric     &  81.5385 &  93.8462 & 95.3846 \\ 
     Log     & 84.6154  & 90.7692  & 96.9231 \\
    Cauchy     &86.1538  &   92.3077 &  98.4615  \\
    HI   & 86.1538& 95.3846 &  96.9231  
  \end{tabular}}
  \end{center}  
  \caption{  This table shows an ablation study;  F-SV, F-$\alpha$  stand respectively for fixed support vectors and fixed mixing  parameters $\alpha$ while   L-SV, L-$\alpha$ stand for learned ones. Note that the results, when learning the support vectors using the  linear kernel, are identical (both with  fixed and learned $\alpha$)  as one may include the multiplicative factors $\alpha$ in the learned support vectors (the converse is not true).}
\label{table3}
\end{table} 

\noindent  Finally, we compare the classification performances   of our KGCN against other related methods in action recognition  ranging from sequence based such as LSTM and GRU \cite{DeepGRU,GCALSTM,STALSTM} to deep graph (non-vectorial) methods based on spatial and spectral convolution \cite{kipf17,SGCCONV19,Bresson16}. From the results in table \ref{compare},  our KGCN brings a substantial gain w.r.t state of the art methods, and provides comparable results with the best vectorial methods.
 \begin{table}[!htb]
  \begin{center}
\resizebox{0.38\linewidth}{!}{
\begin{adjustbox}{angle=-90}
\setlength\tabcolsep{2.4pt}
\def\arraystretch{0.7}  \begin{tabular}{c||ccccccccccccccccccc}
   \rotatebox{90}{Perfs} &     \rotatebox{90}{90.00} &  \rotatebox{90}{96.00} &  \rotatebox{90}{94.00}&  \rotatebox{90}{96.00}&   \rotatebox{90}{49.7 }&  \rotatebox{90}{80.3 }&  \rotatebox{90}{86.9 }&  \rotatebox{90}{83.9 }&  \rotatebox{90}{80.35 }&  \rotatebox{90}{90.41}&   \rotatebox{90}{93.3 } &  \rotatebox{90}{90.5}&   \rotatebox{90}{91.51}&  \rotatebox{90}{94.9}&  \rotatebox{90}{97.2}&  \rotatebox{90}{95.7}&  \rotatebox{90}{93.7 } &  \rotatebox{90}{{{\bf 98.46} } }\\  
     &  &  &  &  &  &  &  &  &  &  &        &  &  &  &  &  &  &  &     \\
     \rotatebox{90}{Methods} &    \rotatebox{90}{ GCNConv \cite{kipf17}} & \rotatebox{90}{ArmaConv \cite{ARMACONV19}} & \rotatebox{90}{ SGCConv \cite{SGCCONV19}} & \rotatebox{90}{ ChebyNet \cite{Bresson16}}& \rotatebox{90}{  Raw coordinates  \cite{SBU12}} & \rotatebox{90}{Joint features \cite{SBU12}} & \rotatebox{90}{Interact Pose \cite{InteractPose}} & \rotatebox{90}{CHARM \cite{CHARM15}} & \rotatebox{90}{ HBRNN-L \cite{HBRNNL15}} & \rotatebox{90}{Co-occurrence LSTM \cite{CoOccurence16}} & \rotatebox{90}{ ST-LSTM \cite{STLSTM16}}  & \rotatebox{90}{ Topological pose ordering\cite{velocity2}} & \rotatebox{90}{ STA-LSTM \cite{STALSTM}} & \rotatebox{90}{ GCA-LSTM \cite{GCALSTM}} & \rotatebox{90}{ VA-LSTM  \cite{VALSTM}} & \rotatebox{90}{DeepGRU  \cite{DeepGRU}} & \rotatebox{90} {Riemannian manifold trajectory\cite{RiemannianManifoldTraject}}  &  \rotatebox{90}{Our best KGCN model}   \\    
 \end{tabular}
\end{adjustbox}
 }
 \caption{ Comparison against state of the art methods.}   \label{compare}            
\end{center}
\end{table}
\section{Conclusion}
We introduce in this paper a novel GCN formulation based on kernel machines. The method defines convolutional graph filters in the span of nodes in a (high or potentially infinite dimensional) reproducing kernel Hilbert space ({\it RKHS}), with the particularity that node representations, in the {\it RKHS}, are learned instead of being taken from training data. This makes the proposed approach (semi-)parametric and tractable while also being effective and less subject to overfitting. Indeed, the proposed GCN formulation is dual and requires few parameters, it also provides an effective way to enhance the discrimination power of the learned graph representations and it overtakes standard (primal) GCN approaches as well as the related work. \\
As a future work, we are currently investigating the combination of explicit node expansion with implicit kernel mapping, in order to further enhance the generalization performances of other pattern recognition tasks.

\end{document}